\documentclass[10pt,twocolumn,letterpaper]{article}

\usepackage{graphicx}
\usepackage{amsmath}
\usepackage{amssymb}
\usepackage{booktabs}
\usepackage{tikz}
\usepackage{pifont}
\newcommand{\xmark}{\ding{55}}%

\usepackage[pagebackref,breaklinks,colorlinks]{hyperref}

\usepackage[capitalize]{cleveref}
\crefname{section}{Sec.}{Secs.}
\Crefname{section}{Section}{Sections}
\Crefname{table}{Table}{Tables}
\crefname{table}{Tab.}{Tabs.}

\begin{document}

\title{Human Vision Based 3D Point Cloud Semantic Segmentation of Large-Scale Outdoor Scene}
\date{\vspace{-5ex}}
\author{Sunghwan Yoo, Yeonjeong Jeong, Maryam Jameela, Gunho Sohn\\
Department of Earth and Space Science and Engineering\\ 
Lassonde School of Engineering York University, Canada\\
{\tt\small (jacobyoo, yjjeong, maryumja, gsohn)@yorku.ca}
}
\maketitle

\begin{abstract}
This paper proposes EyeNet, a novel semantic segmentation network for point clouds that addresses the critical yet often overlooked parameter of coverage area size. Inspired by human peripheral vision, EyeNet overcomes the limitations of conventional networks by introducing a simple but efficient multi-scale input and a parallel processing network with connection blocks between parallel streams. The proposed approach effectively addresses the challenges of dense point clouds, as demonstrated by our ablation studies and state-of-the-art performance on Large-Scale Outdoor datasets.
\end{abstract}

\section{Introduction}
\label{introduction}

Recently, there has been growing interest in developing digital twins of the three-dimensional world, driven by their various applications. With advancements in LiDAR devices and survey techniques, point cloud datasets have become more accurate, dense, and spatially extensive, both on the ground level\cite{paris2014,iqmulus,hackel2017semantic3d,semantickitti,toronto3d} and in the airborne level\cite{isprs2012, dales, sensaturban, sum}. \par

However, the functional coverage area of an input batch is critical for effective feature learning in semantic segmentation networks, especially as data sets become denser\cite{sensaturban, dales}. Traditional approaches to semantic segmentation, such as increasing the number of input points\cite{randla,kpconv,pointtransformers} and downsampling the point cloud\cite{PointNet, pointnet++} to extract features from high-density outdoor scenes, have limitations. For instance, increasing the number of input points is restricted by the memory capacity of GPUs, while further downsampling can result in information loss.

\begin{figure}[ht]
  \centering
  \includegraphics[trim=5cm 4cm 7cm 5cm, clip, width=\linewidth]{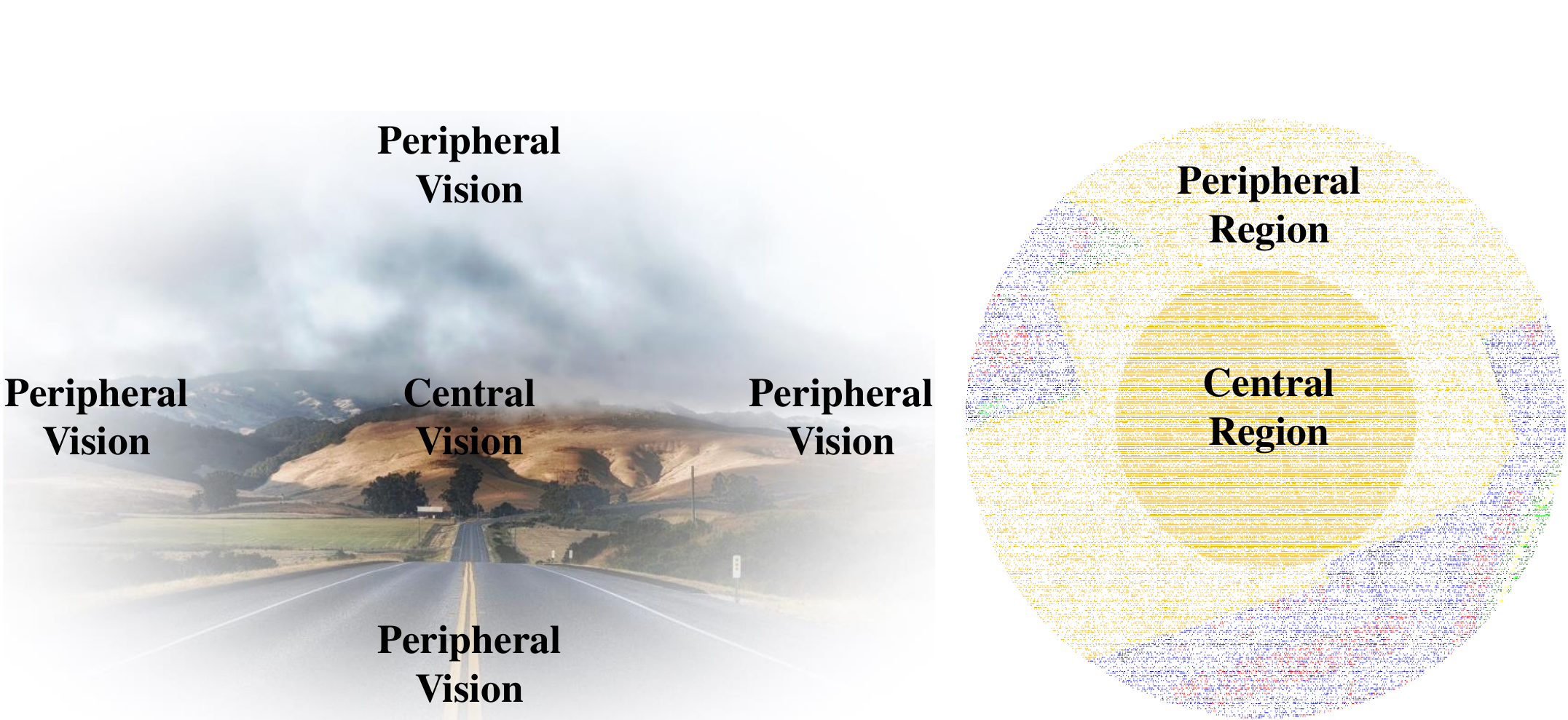}
  \caption{The design of EyeNet is based on the concept of the human vision system. We drew inspiration from the peripheral vision system, which has a low density of receptors but covers a larger region than the central vision. As a result, we added the peripheral region around the central region to achieve a similar effect in our network.}
  \label{fig:concept}
\end{figure}

In this work, we propose a 3D semantic segmentation network called EyeNet, inspired by the human visual system, peripheral vision. The network takes input from multi-scale regions, including the central and peripheral regions. The high-density central region is advantageous for extracting precise geometric features, while the low-density peripheral region with a larger coverage area is beneficial for extracting features from objects that require such assets. To process these regions, the study uses a parallel feature processing design with connection layers for exchanging features between parallel streams. The proposed network outperforms state-of-the-art semantic segmentation methods on the SensatUrban benchmark\cite{sensaturban} and Toronto3D\cite{toronto3d}. The design philosophy of the EyeNet is presented in \Cref{fig:concept}.

Our key contributions are:
\begin{itemize}
    \item Our approach introduces a memory-efficient human vision based input that draws inspiration from human vision, in order to address the limitations of traditional semantic segmentation networks.
    \item To facilitate feature learning using the human vision based input, we proposed an efficient parallel processing architecture.
    \item To ensure that complex features from both the central and peripheral regions are preserved and effectively merged for improved learning, we proposed an effective deep learning connection layer.
\end{itemize}

\begin{figure*}
  \centering
  \includegraphics[trim=3cm 5cm 2cm 5cm,width=\linewidth, clip]{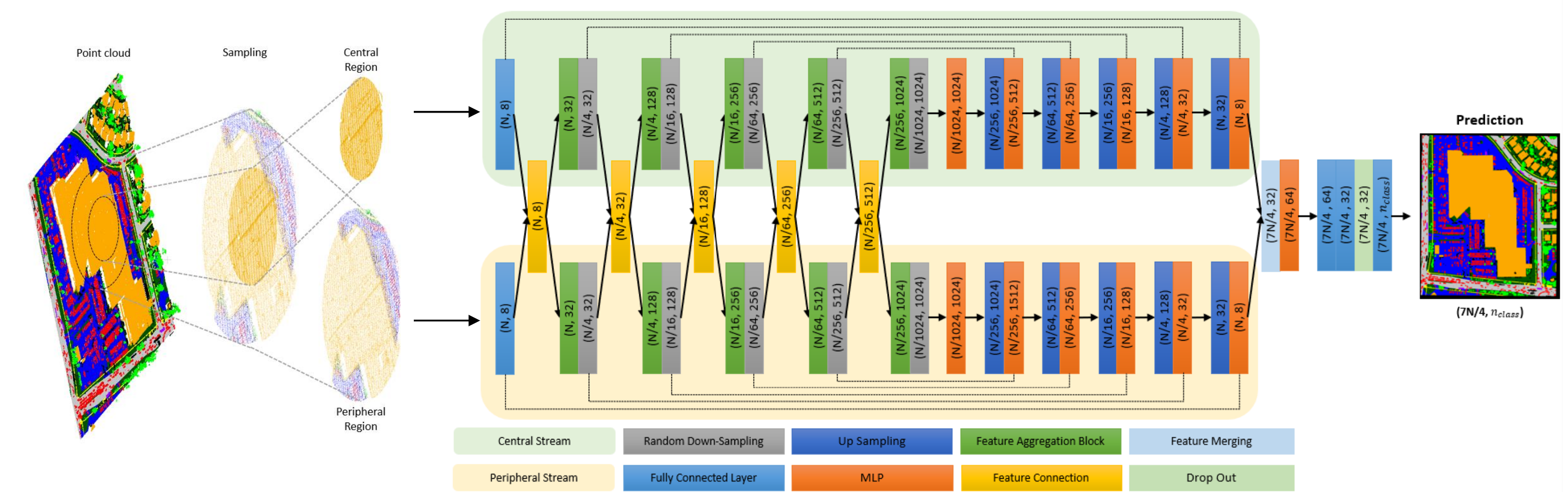}
  \caption{The figure displays the architecture of EyeNet, which utilizes a parallel processing approach to handle the multi-contour input. The network also includes connection and feature merging blocks to facilitate information exchange between the central and peripheral streams. N and d are used to represent the number of points and feature dimension, respectively. The central and peripheral regions share $N/4$ points as inputs, resulting in a total of $7N/4$ input points. MLP represents the multilayer perception.}
  \label{fig:architecture}
\end{figure*}

\section{Related Work}
Semantic segmentation of outdoor 3D point clouds is a challenging task that has received increasing attention recently. Numerous deep learning-based approaches have been proposed to address this problem.\cite{PointNet, pointnet++, sparseconv,gacnet,randla,kpconv,pointtransformers} \par

Deep learning-based methods have shown promising results in semantic segmentation of outdoor 3D point clouds. However, achieving optimal coverage area of the deep learning network is one of crucial factors. Two commonly used methods to achieve an effective coverage area in outdoor 3D point clouds are downsampling the input point cloud and increasing the number of input points.

\subsection{Downsampling Input Point Cloud}
\label{DownsamplingInputPointCloud}
One of the most effective ways to manage computational costs and ensure sufficient coverage area of the input in outdoor 3D point cloud semantic segmentation is downsampling. Early networks, such as PointNet\cite{PointNet,pointnet++} often utilized the constant-volume downsampling method, which collects points in a fixed volume and downsamples them to a fixed number of points. More recent networks such as KPConv\cite{kpconv, randla} adopted the grid sampling method where one point per voxel in a 3D grid was kept.  Although this approach is computationally efficient, it disregards differences in point distribution within the point cloud, resulting in a loss of geometric information and a decrease in performance\cite{sensaturban}.

\subsection{Increasing the Number of Inputs}
\label{IncreasingtheNumberofInputs}
Recently, the trend has shifted away from constant-volume downsampling towards increasing the number of input points through grid sampling\cite{randla, randlanet++, kpconv, pointtransformers}. RandLA\cite{randla} is a network that processes large-scale point clouds by utilizing random sampling based on k-nearest neighbors (KNN), thus avoiding the loss of geometric information. However, a critical limitation of this method is that the ability to increase the number of input points can be restricted by the memory capacity of GPUs.

\subsection{Peripheral Vision in Computer Vision}
\label{PeripheralVisioninComputerVision}
The integration of peripheral vision into deep learning has been a widely explored area in 2D computer vision research. It is primarily investigated in 2D object detection and 2D semantic segmentation\cite{peripheralml1, peripheralml2, peripheralml3, peripheralml4}. Also, recent studies have explored the combination of peripheral vision and transformer networks\cite{peripheraltransformer}, leading to the development of the Peripheral Vision Transformer network. However, the application of peripheral vision systems in 3D computer vision tasks has not been extensively explored.

\setlength{\belowdisplayskip}{0pt} \setlength{\belowdisplayshortskip}{0pt}
\setlength{\abovedisplayskip}{0pt} \setlength{\abovedisplayshortskip}{0pt}
\newcommand*\circled[1]{\tikz[baseline=(char.base)]{
  \node[shape=circle,draw,inner sep=1pt] (char) {#1};}}
  
\section{Methodology}
\label{Methodology}
\subsection{Overview}
\label{Overeview}
Real-world 3D scene's point clouds have tremendous points and varying densities, requiring sampling methods like grid subsampling which scale is determined by voxel size\cite{thomasthesis}. Then, unsampled points are predicted based on the nearest sampled point’s prediction usually decided by voting because random cropping is used due to GPU memory limit.  When we crop the input point cloud randomly for training and inference, the optimal coverage size of the input point cloud is not well-defined.\par
To address this issue, we introduce a novel approach in EyeNet that utilizes both central and peripheral regions, allowing for a parallel processing network that facilitates communication between these regions to retain important features. By incorporating the human vision based input in this manner, we can maximize the benefits of these inputs. A summary of the EyeNet architecture is shown in \Cref{fig:architecture}. Our Eyenet can basically be applied to most existing point-based networks\cite{randla,kpconv,pointtransformers} easily. In this paper, we select RandLA as our baseline. Therefore, any process not described in this section followed the same procedures used in the baseline RandLA implementation\cite{randlagithub}.\par
As depicted in \Cref{fig:architecture}, we constructed the network by stacking two streams of RandLA-Net's encoder and decoder in parallel. To extract fundamental features in each encoding layer, we utilized the k-NN based RandLA's local feature aggregation (LFA) module. The LFA module outputs from both streams were randomly downsampled and fused through the connection layer. After the five encoding layers, we included five decoding layers with a bilinear interpolation operation. Skip connections were utilized to concatenate features from encoding and decoding layers to minimize information loss due to random downsampling. Finally, the outputs from both streams were merged using the Feature Merging Block and passed through fully connected layers (fc) to assign predicted labels to each point.

\subsection{Human Vision Based Input}
\label{PeripheralInput}
Due to the extensive number of points present in urban scale large outdoor point cloud, it is not feasible to input all points into the network simultaneously. Thus, for both training and inference purposes, it is necessary to crop the input region in the entire scene. To do this, our approach, called Human Vision Based Input, is centred around two key factors: the coverage areas and point densities of the central and peripheral regions. Let N be the number of input points for each feature processing stream. We use the K-nearest neighbours (KNN) algorithm to gather N points from the centroid $C_0$, forming the input points for the central stream $P_c= \{ p_c^1 \cdots p_c^n \cdots p_c^N \}$. Here, we define $R$ as the distance between $C_0$ and the farthest point $p_c^f$ in $P_c$, and we set the radius of the central region to $R$ and the peripheral region to $2R$.

We randomly sample $N/4$ points from the central region, and we refer to these points as the messenger points, which are represented as $P_m= \{ p_m^1 \cdots p_m^n \cdots p_m^{N/4} \} \subseteq P_c$.

Moreover, we randomly select an additional $3N/4$ points from the region between $R$ and $2R$, and we define them as $P_{ph-m}= \{ p_{ph}^1 \cdots p_{ph}^n \cdots p_{ph}^{3N/4} \}$. We choose the number $3N/4$ to ensure uniform point density with $P_m$.

\begin{figure}[ht]
  \centering
  \includegraphics[trim=5.5cm 5cm 5.5cm 5cm,width=\linewidth, clip]{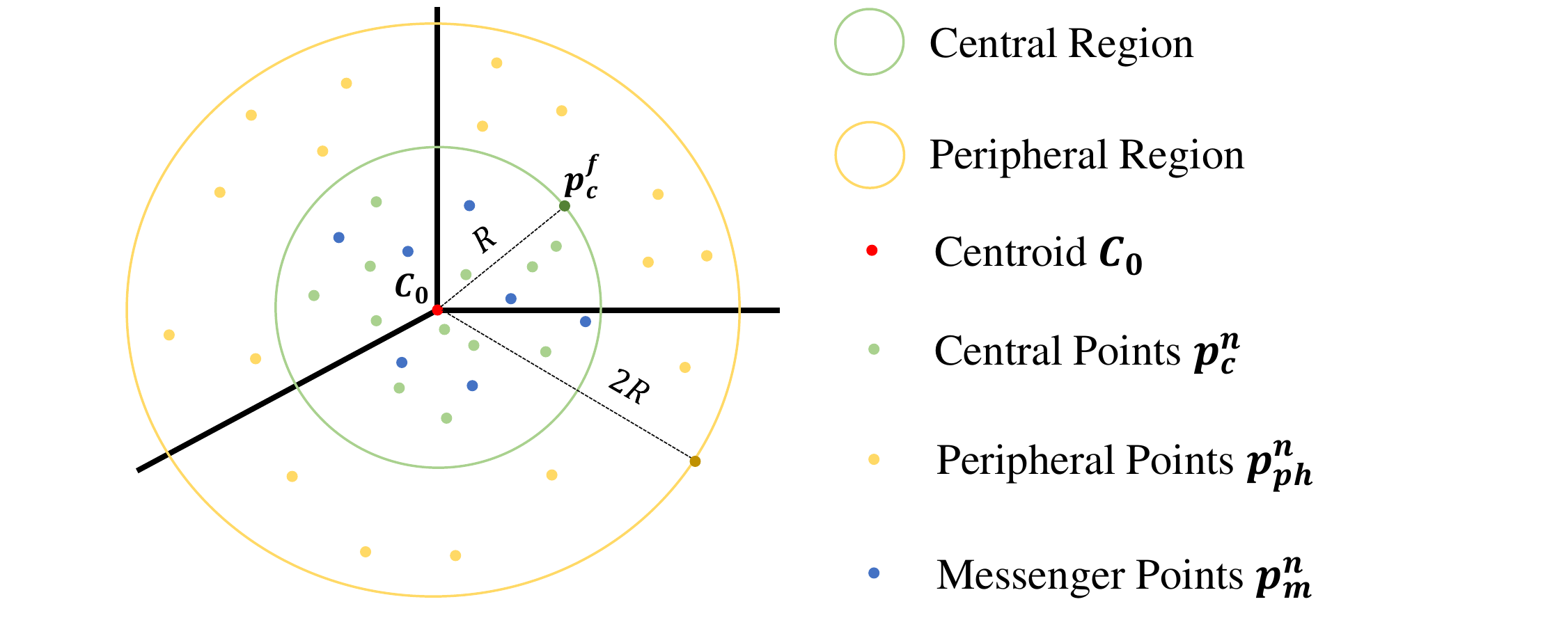}
  \caption{The overview of our human vision based input. Points in the central region are generated based on number-based sampling, and points in the peripheral region are generated based on volume-based down sampling.}
   \label{fig:PeripheralInput}
\end{figure}

Finally, the combination of the point clouds $P_{ph-m}$ and $P_m$ becomes the input point cloud for the peripheral stream, which we denote as $P_{ph} = P_{ph-m} + P_m   = \{ p_{ph}^1 \cdots p_{ph}^n \cdots p_{ph}^{3N/4}, p_m^1 \cdots p_m^n \cdots p_m^{N/4} \}$. Therefore, the messenger points are an intersection between the points in the central and peripheral regions. An overview of the operation is shown in \Cref{fig:PeripheralInput}.

Our proposed human vision based technique for collecting network input points allows us to cover the same input region for a deep neural network using substantially fewer points than the conventional fixed-number-based approaches like RandLA. In all experiments, we used nearly 60\% fewer points than the baseline. The efficient sampling approach enabled our EyeNet to utilize GPU memory more effectively. During inference, our EyeNet was capable of employing four times more batches than the baseline due to this approach.
\begin{figure*}[t]
  \centering
  \includegraphics[trim=0.5cm 0cm 0.5cm 0cm,width=\linewidth, clip]{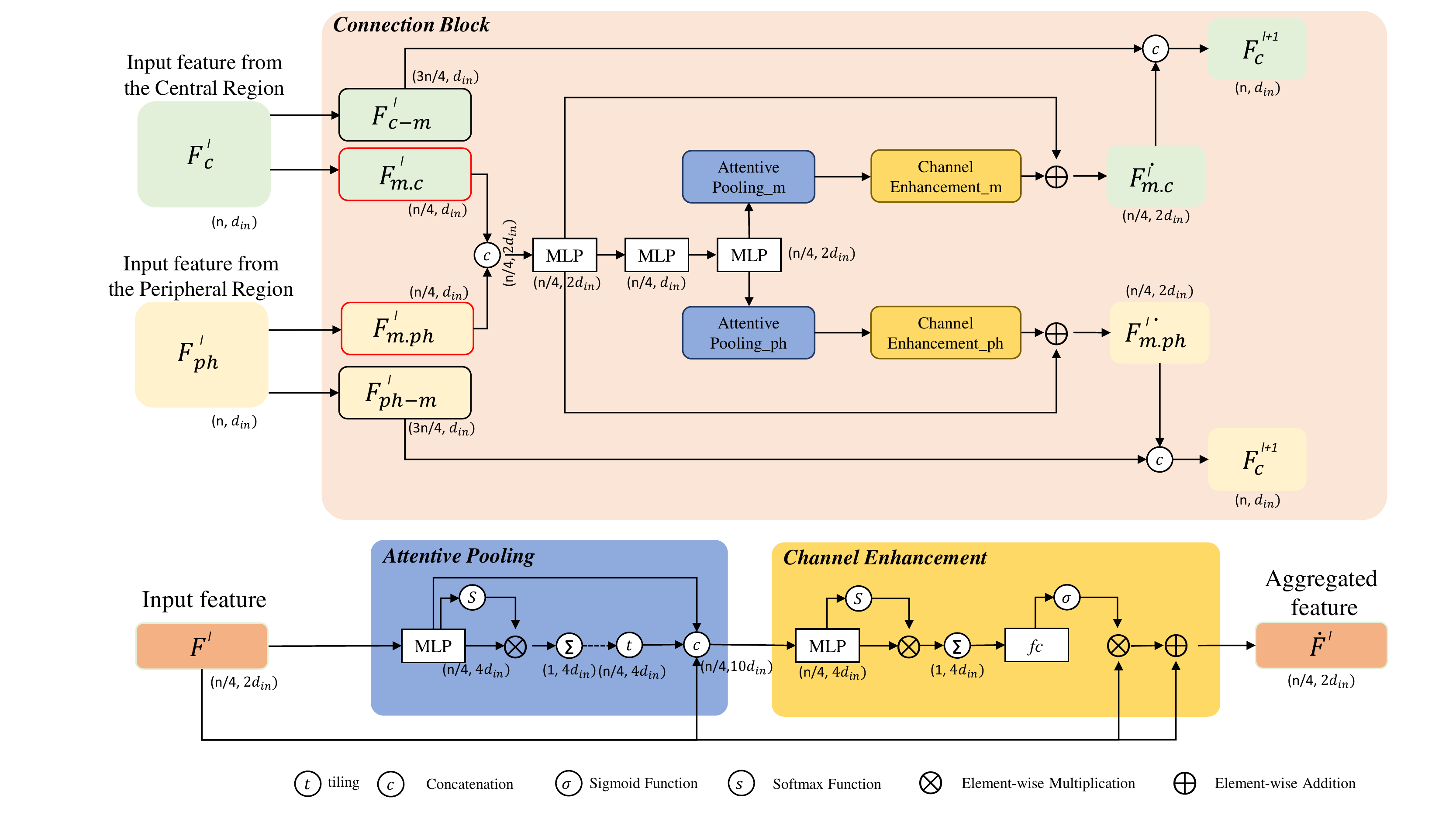}
  \caption{The proposed connection block. The top panel shows the overview of the connection block that is developed to facilitate information exchange between the central and peripheral streams. The bottom panel shows the self-attentive pooling block that weighs the most important features and the channel enhancement block that further enhances the features from the attentive pooling block. $fc$: the fully connected layer.}
   \label{fig:overviewofmrc}
\end{figure*}
\begin{figure*}[t]
  \centering
  \includegraphics[trim=3cm 5cm 3cm 5cm,width=\linewidth, clip]{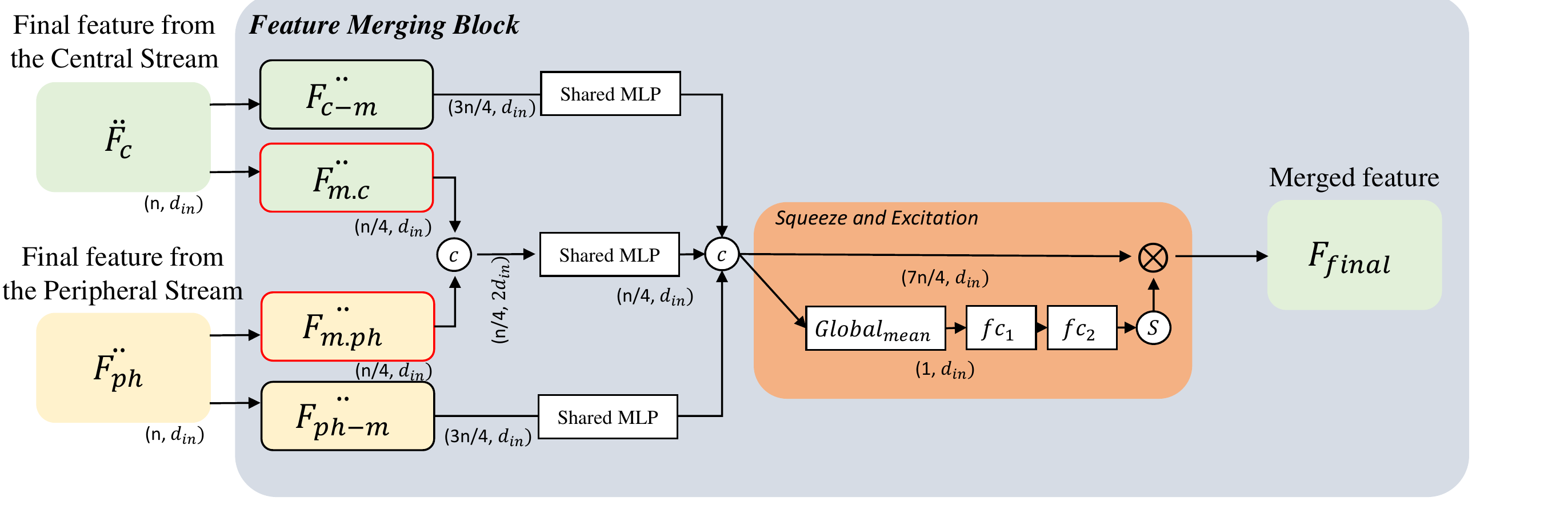}
  \caption{The overview of our feature merging block.  First, the features are processed, and shared MLPs are applied to smooth the feature spaces. Then, the point-wise concatenation of the features occurs, followed by the squeeze and excitation that weighs the most important features for predictions.}
  \label{fig:FeatureMerging}
\end{figure*}
\subsection{Parallel Processing and Receptive Field}
\label{ParallelProcessingForDualReceptiveField}
The network comprises two streams, namely the central and peripheral streams. The central and peripheral streams process point clouds $P_c$ and $P_{ph}$, respectively, with different densities. It makes the receptive field of the peripheral region larger than that of the central region when we use the KNN-based module for feature aggregation with the same number of K for both regions. In this concept, a point can learn its feature from a variable size of the receptive field - small when in the central region, large when in the peripheral region, or both when selected as a messenger point due to the use of random cropping in large-scale 3D point cloud training and inferencing.

\subsection{Connection Block}
\label{Connection}
As we utilize a parallel processing network, establishing connections between two streams is crucial. We were motivated by previous works such as \cite{epnet,3dcvf} and developed the connection block that exchanges information between streams through messenger points. Since the feature information from the central and peripheral streams vary, we adopted the two-way self-attentive pooling block from \cite{randla} along with the channel enhancement block from \cite{randlanet++} for better information exchange.\par

\noindent\textbf{Feature Preparation:} To start, we consider the features of points in the output of the central and peripheral streams, at a given layer $l$, denoted as 
$F_{c}^l = \{ f_{c}^{l\, 1} \cdots f_{c}^{l\, n} \cdots f_{c}^{l\, N^l} \}$ and 
$F_{ph}^l = \{ f_{ph}^{l\, 1} \cdots f_{ph}^{l\, n} \cdots f_{ph}^{l\, N^l} \}$, respectively. From these features, we extract the features of messenger points 
$F_{m.c}^l = \{ f_{m.c}^{l\, 1} \cdots f_{m.c}^{l\, n} \cdots f_{m.c}^{l\, N^l/4} \} \subseteq F_c^l$ and 
$F_{m.ph}^l = \{ f_{m.ph}^{l\, 1} \cdots f_{m.ph}^{l\, n} \cdots f_{m.ph}^{l\, N^l/4} \} \subseteq F_{ph}^l$. The remaining points' features are referred to as 
$F_{c-m}^l = \{ f_{c-m}^{l\, 1} \cdots f_{c-m}^{l\, n} \cdots f_{c-m}^{l\, 3N^l/4} \} \subseteq F_c^l$ and
$F_{ph-m}^l = \{ f_{ph-m}^{l\, 1} \cdots f_{ph-m}^{l\, n} \cdots f_{ph-m}^{l\, 3N^l/4} \} \subseteq F_{ph}^l$.
Afterward, we concatenate the channel-wise features of $F_{m.c}^l$ and $F_{m.ph}^l$ into $F_m^l$and process the resulting output through shared MLP function $m()$ three times. The shared MLPs' operations are denoted as follows:
\begin{equation}
    \begin{split}
        f^{l\, i} =m_3 \left ( m_2 \left ( m_1 \left ( f_{m}^{l\, i} ,W_1\right),W_2\right),W_3\right)
    \end{split}
    \label{eq:F}
\end{equation} where $F^l = \{ f^{l\, 1} \cdots f^{l\, n} \cdots f^{l\, N^l/4} \}$ and $W_s$ are the learnable weights of shared MLPs.

\par

\noindent\textbf{Attentive Pooling:} Inspired by self-attentive pooling block in RandLA\cite{randla}, we used the powerful attention mechanism to exchange and fuse important features automatically, and it consists of the following steps. \par
We utilized a shared MLP followed by softmax to learn a unique attention score for each feature from $F^l$. It is formally defined as follows:
\begin{equation}
    \begin{split}
    f_{ap}^{l\,i} = L\left(m_a\left( f^{l\, i},W_a \right)\right)
     \end{split}
    \label{eq:AP1-1}
\end{equation}
where $W$ is the learnable weights of shared MLP and $L$ is the Relu activation function.
\begin{equation}
    \begin{split}
    s_{ap}^{l\,i} = S\left( f_{ap}^{l\,i} \right)
     \end{split}
    \label{eq:AP1-2}
\end{equation} where $S$ is softmax function. We can consider the attention scores that have been learned as a softmax function that automatically picks out significant characteristics. In technical terms, these traits are combined through a weighted summation process, as described below:
\begin{equation}
    \begin{split}
    {f^{l\,i}}' = \sum \left( f_{ap}^{l\, i} \cdot  s_{ap}^{l\,i}\right)
    \end{split}
    \label{eq:AP2}
\end{equation}\par

\noindent\textbf{Channel Enhancement:} In addition, we improved the features by creating more prominent global features while minimizing the loss of information during the self-attention encoding process. This method was initially presented in RandLA++\cite{randlanet++} and involved the following procedures.\par

To start, intermediate feature sets $F^{l}$, ${F^{l}}'$, and $F^{l}_{ap}$ are concatenated into $F_{cat}^{l}$. Later, the concatenated features are operated by shared MLP function $m_{ce}()$, followed by ReLu $L$.
\begin{equation}
    \begin{split}
    f^{l\,i}_{ce} = L\left (m_{ce}\left (f^{l \, i}_{cat}, W_{ce}  \right )  \right )
    \end{split}
    \label{eq:CE1}
\end{equation} 
Subsequently, the output feature $f^{l\,i}_{ce}$ is improved channel-wise by attention pooling using \cref{eq:CE2_1} and \cref{eq:CE2_2} into $f^{l\,i}_{ceat}$. Ultimately, a fully connected layer yields the combined feature $\dot{f^{l\,i}}$. It is formally defined as follows:
\begin{equation}
    \begin{split}
    {s^{l\,i}}_{ce} = S\left( f^{l\,i}_{ce} \right)
    \end{split}
    \label{eq:CE2_1}
\end{equation}

\begin{equation}
    \begin{split}
    {f^{l\,i}}_{ceat} = \sum \left( f^{l\,i}_{ce} \cdot  {s^{l\,i}}_{ce}\right)
    \end{split}
    \label{eq:CE2_2}
\end{equation}

\begin{equation}
    \begin{split}
    \dot{f^{l\,i}}_{ce} = \sigma \left(fc\left({f^{l\,i}}_{ceat}, W_{fc}\right)\right)
    \end{split}
    \label{eq:CE2_3}
\end{equation} where $\sigma$ is a sigmoid function. \par
Finally, element-wise multiplication between $F^l$ and $\dot{F_{ce}^l}$ is applied, which is then added to $F^l$ to generate channel enchanced feature $\dot{F^l}$.\par
Lastly, the outputs of both branches, $\dot{f_c^{l \, i}}$ and $\dot{f_{ph}^{l \, i}}$ respectively, are added with the corresponding output of $m_1$ operation of the concatenated messenger features. Then point-wise concatenation is performed with the corresponding remaining points' features. It is defined as follows:
\begin{equation}
    \begin{split}
    \dot{f_{m.c}^{l \, i}} = \dot{f_c^{l \, i}} + m_1 \left( f_{m}^{l\, i} ,W_1\right)
    \end{split}
    \label{eq:fmc}
\end{equation}

\begin{equation}
    \begin{split}
    \dot{f_{m.ph}^{l \, i}} = \dot{f_{ph}^{l \, i}} + m_1 \left( f_{m}^{l\, i} ,W_1\right)
    \end{split}
    \label{eq:fmph}
\end{equation}

In summary, we begin by preparing features through our feature preparation process. These features are then divided into two streams of attention pooling and channel enhancing blocks. This is done to allow for the effective exchange and merging of features needed for both the central and peripheral streams. Please see the diagram of our connection block in Figure \ref{fig:overviewofmrc} for further illustration.

\subsection{Feature Merging}
\label{FeatureMerging}
The output features $\ddot{F_c}$ and $\ddot{F_{ph}}$ from the central and peripheral streams, respecetively, have to be merged back to match the input size of the human vision based input. We utilize squeeze and excitation method\cite{squeeze_excitation} to perform feature merging. \par

First, we extract $\ddot{F_{m.c}}$ and $\ddot{F_{m.ph}}$, features of messenger points, from $\ddot{F_c}$ and $\ddot{F_{ph}}$ respectively, resulting in corresponding remained features $\ddot{F_{c-m}}$ and $\ddot{F_{ph-m}}$. Next, we perform channel-wise concatenation between $\ddot{F_{m.c}}$ and $\ddot{F_{m.ph}}$ to obtain $\ddot{F_{m}}$. We then use shared MLPs to smooth the feature spaces of $\ddot{F_{m}}$, $\ddot{F_{c-m}}$, and $\ddot{F_{ph-m}}$, followed by point-wise concatenation. Subsequently, we apply a squeeze and excitation module to enhance the salient features. The feature merging operation is illustrated in \Cref{fig:FeatureMerging}.\par

\def\rot{\rotatebox}

\begin{table*}
  \centering
  \resizebox{\textwidth}{!} {
  \begin{tabular} {r c c c c c c c c c c c c c c c}
    \toprule
    Method &  \rot{90}{OA(\%)} & \rot{90}{mIoU(\%)} & \rot{90}{ground} & \rot{90}{veg.} & \rot{90}{building} & \rot{90}{wall} & \rot{90}{bridge} & \rot{90}{parking} & \rot{90}{rail} & \rot{90}{traffic} & \rot{90}{street} & \rot{90}{car} & \rot{90}{footpath} & \rot{90}{bike} & \rot{90}{water} \\
    \hline
    PointNet\cite{PointNet} & 
    80.78 & 23.71 &
    67.96 & 89.52 & 80.05 & 0.00 & 0.00 & 3.95 & 0.00 & 31.55 & 0.00 & 35.14 & 0.00 & 0.00 & 0.00 \\
    
    PointNet++\cite{pointnet++} &
    84.30 & 32.92 &
    72.46 & 94.24 & 84.77 & 2.72 & 2.09 & 25.79 & 0.00 & 31.54 & 11.42 & 38.84 & 7.12 & 0.00 & 56.93 \\

    TangentConv\cite{tangentconv} &
    76.97 & 33.30 &
    71.54 & 91.38 & 75.90 & 35.22 & 0.00 & 45.34 & 0.00 & 26.69 & 19.24 & 67.58 & 0.01 & 0.00 & 0.00\\

    SPGraph\cite{spgraph} &
    85.27 & 37.29 &
    69.93 & 94.55 & 88.87 & 32.83 & 12.58 & 15.77 & 15.48 & 30.63 & 22.96 & 56.42 & 0.54 & 0.00 & 44.24 \\

    SparseConv\cite{sparseconv} &
    88.66 & 42.66 &
    74.10 & 97.90 & 94.20 & 63.30 & 7.50 & 24.20 & 0.00 & 30.10 & 34.00 & 74.40 & 0.00 & 0.00 & 54.80 \\

    KPConv\cite{kpconv} &
    93.20 & 57.58 &
    \textbf{87.10} & 98.31 & 95.33 & \textbf{74.40} & 28.69 & 41.38 & 0.00 & 55.99 & \textbf{54.43} & \textbf{85.67} & 40.39 & 0.00 & \textbf{86.30} \\

    RandLA\cite{randla} &
    89.78 & 52.69 &
    80.11 & 98.07 & 91.58 & 48.88 & 40.75 & 51.62 & 0.00 & 56.67 & 33.23 & 80.14 & 32.63 & 0.00 & 71.31 \\

    RandLA++\cite{randlanet++} &
    91.90 & 57.10 &
    84.10 & 98.20 & 94.80 & 58.60 & \textbf{59.80} & 53.40 & 0.00 & 54.60 & 42.60 & 78.20 & 38.20 & 0.00 & 69.70 \\
    \hline

    EyeNet(ours) &
    \textbf{93.70} & \textbf{62.30} &
    86.60 & \textbf{98.60} & \textbf{96.20} & 65.80 & 59.20 & \textbf{64.80} & \textbf{17.90} & \textbf{64.80} & 49.80 & 83.10 & \textbf{46.20} & \textbf{11.10} & 65.40 \\
    
    \bottomrule
  \end{tabular}
  }
  \caption{Sensat Urban Performance Comparison. Performances of other works except RandLA++ are taken from the SensatUrban paper\cite{sensaturban} RandLA++ performance is taken from its paper. \cite{randlanet++} }
  \label{tab:SensatUrbanPerformanceComparsion}
\end{table*}

\begin{table*}
  \centering
  \resizebox{\textwidth}{!} {
  \begin{tabular} {r c c c c c c c c c c}
    \toprule
    Method &  {OA(\%)} & {mIoU(\%)} & {road} & {road mark.} & {natural} & {building} & {utility line} & {pole} & {car} & {fence} \\
    \hline
    PointNet++\cite{pointnet++} & 
    92.56 &	59.47 &	92.90 & 0.00 & 86.13 & 82.15 & 60.96 & 62.81 & 76.41 & 14.43 \\
    
    DGCNN\cite{dgcnn} &
    94.24&61.79	&93.88	&0.00	&91.25&	80.39&	62.40&	62.32&	88.26&	15.81 \\

    KPConv\cite{kpconv} &
    95.39&	69.11&	94.62&	0.06&	96.07&	91.51&	 \textbf{87.68}&	81.56&	85.66&	15.72\\

    MS-PCNN\cite{ms_pcnn} &
    90.03&	65.89&	93.84&	3.83&	93.46&	82.59&	67.80&	71.95&	91.12&	22.50 \\

    TGNet\cite{tgnet} &
    94.08&	61.34&	93.54&	0.00&	90.83&	81.57&	65.26&	62.98&	88.73&	7.85 \\

    MS-TGNet\cite{toronto3d} &
    \textbf{95.71}&	70.50&	94.41&	17.19&	95.72&	88.83&	76.01&	73.97&	\textbf{94.24}&	23.64 \\

    RandLA\cite{randla} &
    92.95&	77.71&	94.61&	42.62&	96.89&	93.01&	86.51&	78.07&	92.85&	37.12 \\

    Rim et al.\cite{rimetal} &
    72.55&	66.87&	92.74&	14.75&	88.66&	\textbf{93.52}&	81.03&	67.71&	39.65&	\textbf{56.90} \\

    MappingConvSeg\cite{mappingconvseg} &
    93.17&	77.57&	95.02&	39.27&	96.77&	93.32&	86.37&	79.11&	89.81&	40.89 \\
    \hline

    EyeNet(ours) &
    94.63&	\textbf{81.13}&	\textbf{96.98}&	\textbf{65.02}&	\textbf{97.83}&	93.51&	86.77&	\textbf{84.86}&	94.02&	30.01 \\
    
    \bottomrule
  \end{tabular}
  }
  \caption{Toronto3D Performance Comparison. RGB features are not used to obtain results. Performances of other works are taken from the Toronto3D paper\cite{toronto3d}}
  \label{tab:Toronto3DComparsion}
\end{table*}

\begin{table*}
  \centering
  \resizebox{\textwidth}{!} {
  \begin{tabular} {r c c c c c c c c c c}
    \toprule
    Method &  {OA(\%)} & {mIoU(\%)} & {ground} & {vegatation} & {car} & {truck} & {power lines} & {poles} & {fences} & {building} \\
    \hline
    KPConv\cite{kpconv} & 
    \textbf{97.8}&	\textbf{81.1}&	97.1&	\textbf{96.6}&	\textbf{85.3}&	\textbf{41.9}&	75.0&	\textbf{95.5}&	63.5&	94.1 \\
    
    RandLA\cite{randla} &
    97.09&	79.27&	96.99&	93.16&	82.22&	38.6&	\textbf{94.96}&	58.01&	\textbf{73.56}&	96.64 \\

    PointNet++\cite{pointnet++} &
    95.7&	68.3&	94.1&	89.1&	75.4&	30.3&	40.0&	79.9&	46.2&	91.2\\

    ConvPoint\cite{convpoint} &
    97.2&	67.4&	96.9&	96.3&	75.5&	21.7&	40.3&	86.7&	29.6&	91.9 \\

    SPGraph\cite{spgraph} &
    95.5&	60.6&	94.7&	93.4&	62.9&	18.7&	28.5&	65.2&	33.6&	87.9 \\

    PointCNN\cite{toronto3d} &
    97.2&	58.4&	\textbf{97.5}&	95.7&	40.6&	4.8&	57.6&	26.7&	52.6&	91.7 \\

    ShellNet\cite{randla} &
    96.4&	57.4&	96.0&	95.4&	32.2&	39.6&	20.0&	27.4&	60.0&	88.4 \\
    \hline

    EyeNet(ours) &
    97.18&	79.6&	97.15&	93.37&	83.53&	41.3&	94.65&	57.38&	72.54&	\textbf{96.87} \\
    
    \bottomrule
  \end{tabular}
  }
  \caption{DALES Performance Comparison. RandLA results are directly obtained from the author of RandLA. All other results are taken from DALES paper. \cite{dales} }
  \label{tab:DALESComparsion}
\end{table*}

\begin{table*}
  \centering
  \resizebox{\textwidth}{!} {
  \begin{tabular} {r c c c c c c c c c c}
    \toprule
    Method &  {mIoU(\%)}  & {ground} & {vegatation} & {building} & {water} & {car} & {truck} & {traffic road} & {sidewalk} &{parking}\\
    \hline
    RandLA\cite{randla} & 
    58.37&	80.61&	94.44&	95.39&	3.34&	74.59&	13.87&	78.10&	23.43&	61.56 \\
    
    KPConv\cite{kpconv} &
    56.14&	\textbf{86.94}&	\textbf{96.25}&	94.01&	0.00&	\textbf{84.02}&	0.00&	79.93&	3.26&	60.83 \\
    \hline

    EyeNet(ours) &
    \textbf{63.44}&	86.26&	95.94&	\textbf{96.78}&	\textbf{13.61}&	83.02&	\textbf{14.26}&	\textbf{84.65}&	\textbf{31.08}&	\textbf{65.34}\\
    
    \bottomrule
  \end{tabular}
  }
  \caption{YUTO Performance Comparison. Results of RandLA and KPConv are taken from internal experiments.}
  \label{tab:YUTOComparsion}
\end{table*}

\section{Experiments}
\label{Experiments}

\subsection{Implementation details}
\label{Implementationdetails}
We implemented EyeNet in Tensorflow, employing the grid sampling strategy as input pre-processing, followed by KPConv \cite{kpconv} and RandLA \cite{randla}. The Adam optimizer is employed, with an initial learning rate of 0.005, which decays by 5\% after each epoch. For each layer, the number of nearest points K is set to [16, 21, 21, 21, 16]. We sample a fixed number of 28672 points from each point cloud as input, and use a batch size of 16. Our choice of loss function is lovasz-softmax loss \cite{lovasz}. We followed the same voxel size and input features with ones used in RandLA \cite{randla} except YUTO dataset. All experiments are performed on an NVIDIA Quadro RTX 6000 GPU.

\subsection{Semantic Segmentation on Benchmark datasets}
To evaluate the performance of EyeNet, we conducted experiments on the semantic segmentation task. We selected a range of outdoor benchmark datasets, including SensatUrban \cite{sensaturban}, Toronto3D \cite{toronto3d}, DALES \cite{dales}, as well as our own YUTO dataset.

\noindent\textbf{Sensat Urban:} The SensatUrban\cite{sensaturban} dataset for the semantic segmentation task consists of large areas of three UK cities with a coverage area of 7.64 $km^{2}$. 2847 million points were collected using a UAV Photogrammetry and have an average point density of 372 points per square meter. Each point in this dataset is assigned a semantic label of 13 classes (ground, vegetation, building, etc.). The point clouds are divided into 34 tiles, and we follow their work's same train, validation, and test split. For evaluation, we upload the test labels on the online test server\cite{evalautionserver} and acquire mean class-wise intersection over union (mIoU), overall point-wise accuracy (OA), and per class IoU. We set the voxel size of the grid sampling to 0.2m, and use the provided 3D coordinates and color information for training and testing our network.\par

\begin{figure}[t]
  \centering
  \includegraphics[trim=5cm 1.2cm 7cm 0cm,width=\linewidth, clip]{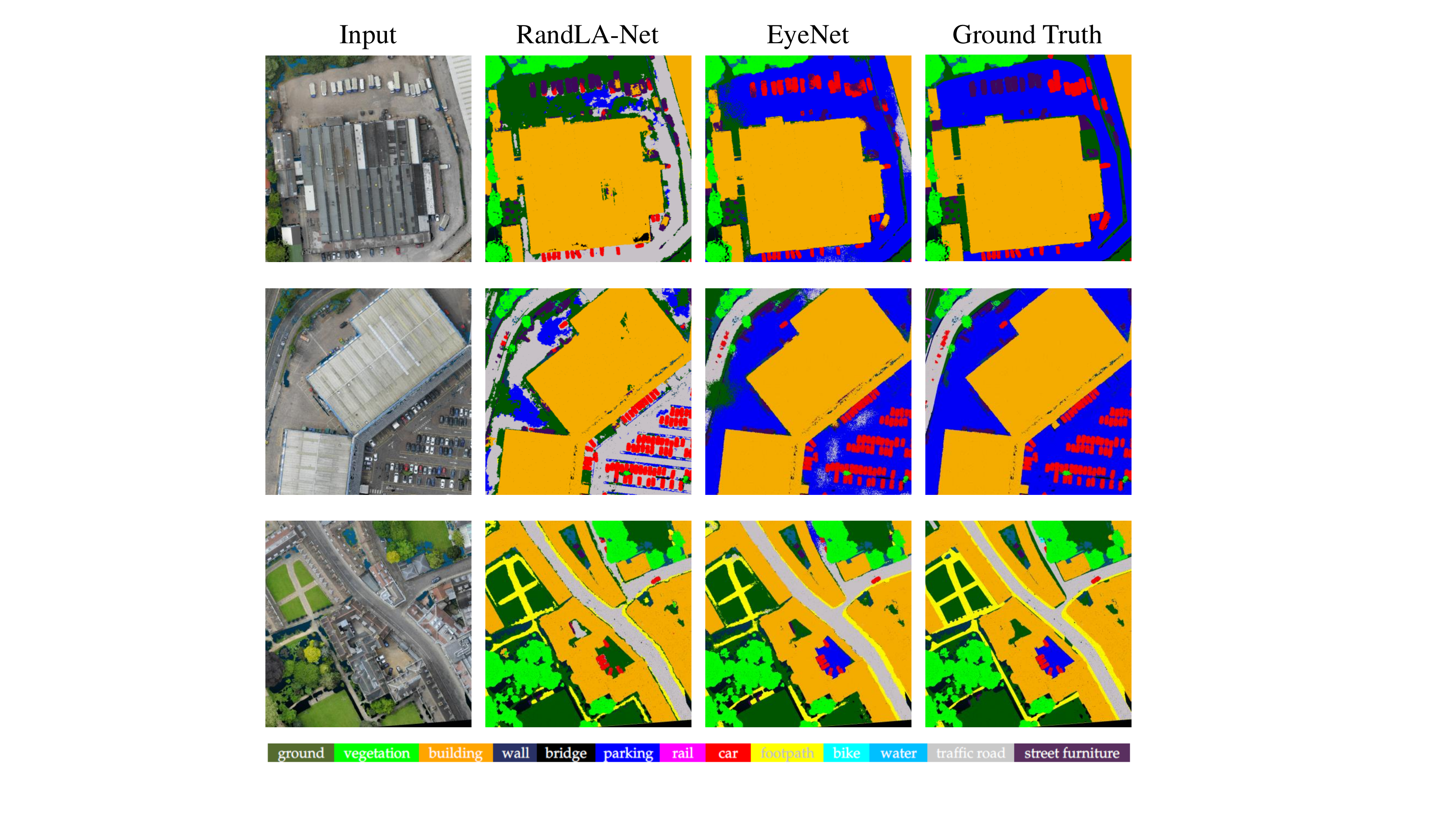}
  \caption{Visualization comparison of SensatUrban between RandLA and EyeNet. From left to right: the input point cloud with RGB colours, semantic segmentation results of RandLA, semantic segmentation results of EyeNet, the ground truth of the input pointcloud}
   \label{fig:sensat_visualization}
\end{figure}

The performance comparison results for SensatUrban are presented in \Cref{tab:SensatUrbanPerformanceComparsion}. EyeNet has set a new state-of-the-art for both OA and mIoU metrics, surpassing all existing methods by a considerable margin. Remarkably, EyeNet has shown the best performance in seven out of thirteen semantic classes. The SensatUrban dataset is notorious for having severe class imbalance, and none of the previous approaches were able to predict the rail and bike classes, except for SPGraph \cite{spgraph}. In contrast, EyeNet successfully predicted these classes without any significant data engineering efforts. Furthermore, recent works such as RandLA \cite{randla}, KPConv \cite{kpconv}, Point Transformers \cite{pointtransformers}, Omni-Supervised \cite{omnisupervised}, FG-Net \cite{fgnet}, and SCF-Net \cite{scfnet} have increased the number of input point clouds up to 100K for expanding the input coverage area. Despite using an input size of approximately 28K, which is far smaller than other approaches, EyeNet demonstrated superior performance in both small and large object classes. An illustration of segmented scenes on Sensat Urban is shown in \cref{fig:sensat_visualization}.\par

\noindent\textbf{Toronto3D:} Toronto3D\cite{toronto3d} is a publicly available dataset that contains 3D point clouds and associated images of urban environments in Toronto, Canada. The dataset covers a total length of 1 kilometer and was collected using a mobile lidar scanner. The dataset contains 78.3 million points with an average point density of 391 points per square meter. Each point in the dataset is labelled with one of 8 semantic classes such as road, road markings, or natural objects. The dataset is divided into four different sections, and the L002 section was selected as the test split for this study. The evaluation was done by calculating mIoU, OA, and per class IoU for the test split. A voxel size of 0.06 meters was used for grid sampling, and the provided 3D coordinates and intensity information were used for training and testing the network.\par

The results of the performance comparison for Toronto3D are presented in Table \ref{tab:Toronto3DComparsion}. The EyeNet model achieved a new state-of-the-art performance for mIoU metric, surpassing all existing methods by a considerable margin. Notably, EyeNet demonstrated the best performance in four out of eight semantic classes. The Toronto3D dataset includes a unique class called road marking, which is challenging to classify without RGB features, and previous approaches have struggled to accurately classify it. However, EyeNet successfully predicted road markings, highlighting its superior performance.\par

\noindent\textbf{DALES:} The Dayton Annotated LiDAR Earth Scan (DALES)\cite{dales} dataset is a publicly available dataset of LiDAR point clouds covering an area of approximately 10 square kilometers in Dayton, Ohio, USA and was collected using an Aerial Laser Scanner (ALS). The dataset includes over 505 million points and has an average point density of 25.5 points per square meter. Each point in the dataset is assigned a semantic label from one of 8 classes, including ground, building, vegetation, water, and others. The dataset is divided into 40 sections and split into training, validation, and test sets for semantic segmentation tasks. The evaluation was done by calculating mIoU, OA, and per class IoU for the test split. A voxel size of 0.32 meters was used for grid sampling, and the provided 3D coordinates was used for training and testing the network.\par

The performance comparison results for DALES are presented in Table \ref{tab:DALESComparsion}. The EyeNet model achieved the second-best performance for mIoU metrics. However, since DALES is the most sparse dataset among the tested benchmarks, the benefits of using the human vision based input in EyeNet are minimal.\par

\noindent\textbf{YUTO:} The York University Toronto Ontario (YUTO) dataset is a privately owned dataset of LiDAR point clouds covering an area of approximately 9.5 square kilometers in the York University campus. The dataset contains 738 million points collected using an ALS and has an average point density of 78 points per square meter. Each point in the dataset is assigned a semantic label from one of nine classes, including ground, traffic road, sidewalk, water, and others. The dataset is divided into 41 tiles, with 11 tiles used for evaluation. The evaluation was performed by calculating mIoU and per-class IoU for the test split. The performance of two popular semantic segmentation networks, RandLA\cite{randla} and KPConv\cite{kpconv}, was also evaluated for comparison. A voxel size of 0.20 meters was used for grid sampling, and the provided 3D coordinates and intensity were used for training and testing the network.\par

The results of the performance comparison for YUTO are presented in Table \ref{tab:YUTOComparsion}. The EyeNet model outperformed the two tested networks by a significant margin in terms of the mIoU metric. It is noteworthy that EyeNet exhibited the highest performance in six out of the nine semantic categories. The YUTO dataset encompasses classes related to the terrain, including traffic road, ground, sidewalk, parking, and water, which is a distinctive feature of ALS datasets. It poses a challenge to classify these classes without the RGB channel accurately. Nevertheless, EyeNet achieved superior results in recognizing these terrain-related classes.\par

\subsection{Ablation Study}
An ablation study was conducted to evaluate the effectiveness of the introduced features in the Sensat Urban test set through an online evaluation server. \par
In order to demonstrate the efficacy of our parallel processing network architecture, we constructed and compared three different types of network structures: Baseline, Sequential, and Parallel. The Baseline structure utilized the pre-existing RandLA-Net architecture. The Sequential structure consisted of two RandLA-Net structures connected sequentially. The Sequential structure was constructed to demonstrate that the performance gain from EyeNet was not solely due to increased parameters. The Parallel structure stacked two RandLA-Net structures in parallel like \cref{fig:architecture}.\par
The study compared the performance of four network structures, including the baseline network RandLA, sequentially stacked RandLAs, EyeNet with only Feature Merging Block, and EyeNet. The results, presented in Table \ref{tab:EffectOnModules}, indicated that the sequential structure resulted in a 6.7\% decrease in mIoU, implying that increasing parameter size alone did not improve performance. In contrast, our EyeNet with only Feature Merging Block yielded improvements of 3.2\% and 8.8\% in OA and mIoU, respectively, over the baseline. These results suggest that our human vision-inspired approach contributed to performance enhancement. Additionally, the connection block led to further improvements of 0.7\% and 0.8\% in OA and mIoU, respectively, indicating that the connection between the two streams improved the network's feature understanding and led to a performance boost.\par

\begin{table}
  \centering
  \renewcommand{\arraystretch}{1.5}
  \begin{tabular} {c c c| c c}
    \toprule
    Peri. Input & Structure & CB &  OA & mIoU \\
    \hline
    \xmark  &  Baseline &\xmark & 89.8 & 52.7 \\
    \xmark & Sequential &\xmark  & 89.4 & 46.0 \\
    \checkmark & Parallel & \xmark  & 93.0 & 61.5 \\
    \checkmark & Parallel & \checkmark & \textbf{93.7} & \textbf{62.3} \\
    
    \bottomrule
  \end{tabular}
  \caption{Ablation results of the peripheral input (Peri. Input),  network structure, and connection block (CB).}
  \label{tab:EffectOnModules}
\end{table}

\section{Conclusion}
\label{conclusion}

Determining an optimal coverage area for the input batch is crucial for effective feature learning in 3D point cloud semantic segmentation, but it is often neglected. To address this issue, we have developed a human vision based input processing semantic segmentation network architecture, which overcomes the limitations of traditional semantic segmentation networks. We have also introduced a merging block, connection block and parallel stack design for effective feature learning. As a result, our approach has achieved state-of-the-art performance on large scale outdoor benchmark datasets such as SensatUrban and Toronto3D. However, it is possible that the performance of our network may be limited by the baseline network. Therefore, in future work, we plan to investigate the application of our method to different network architectures since our method is model-agnostic.

\section{Acknowledgment}
\label{acknowledgment}
This research project has been supported by the Natural Sciences and Engineering Research Council of Canada (NSERC)'s Collaborative Research and Development Grant (CRD) - 3D Mobility Mapping Artificial Intelligence (3DMMAI) and Teledyne Geospatial Inc. We'd like to thank Alvin Poernomo (Machine Learning Developer), Hamdy Elsayed (Innovation Manager) and Chris Verheggen (SVP R\&D).

{\small
\bibliographystyle{ieee_fullname}

}

\end{document}